# 3D Multi-Object Tracking Based on Uncertainty-Guided Data Association

Jiawei He, Chunyun Fu, and Xiyang Wang

*Abstract*—In the existing literature, most 3D multi-object tracking algorithms based on the tracking-by-detection framework employed deterministic tracks and detections for similarity calculation in the data association stage. Namely, the inherent uncertainties existing in tracks and detections are overlooked. In this work, we discard the commonly used deterministic tracks and deterministic detections for data association, instead, we propose to model tracks and detections as random vectors in which uncertainties are taken into account. Then, based on the Jensen-Shannon divergence, the similarity between two multidimensional distributions, i.e. track and detection, is evaluated for data association purposes. Lastly, the level of track uncertainty is incorporated in our cost function design to guide the data association process. Comparative experiments have been conducted on two typical datasets, KITTI and nuScenes, and the results indicated that our proposed method outperformed the compared state-of-the-art 3D tracking algorithms. For the benefit of the community, our code has been made available at https://github.com/hejiawei2023/UG3DMOT.

*Index Terms*—3D multi-object tracking, point clouds, data association.

## I. INTRODUCTION

3D multi-object tracking (MOT) plays an important role in robotics and autonomous driving. Currently, most existing MOT works are based on the typical tracking-by-detection (TBD) paradigm, which decomposes an MOT task into two subtasks: detection and tracking. In a TBD-based MOT method, object detection is first performed and then data association is conducted. In recent years, some impressive 3D object detectors have been proposed in the literature, such as [1]–[6]. These detectors are able to extract 3D information of objects from original point clouds or original images, which greatly facilitates the development of TBD-based MOT technologies.

In many TBD-based tracking methods (e.g., [7]–[9]), track uncertainties are usually taken into account in the track prediction process. For example, Kalman filters have been commonly used to implement this prediction process, since there exists inherent uncertainties in the motion of a tracked object. However, uncertainties are rarely taken into consideration in the data association stages of most existing TBD-based methods. Instead, deterministic tracks and deterministic detections are used for association without considering the uncertainties in tracks and detections [8], [10], [11]. It is well-known that the performance of a TBD-based MOT method hinges on its data association strategy, however, the above methods neglect the inherent uncertainties existing in the tracked objects and the sensor detections, which inevitably reduces the effectiveness and reliability of similarity calculation.

When it comes to tracks, in simple scenarios, it is common to have tracks continuously detected during their life cycle. In this case, track uncertainties are generally small and data association can be completed using deterministic track states without considering their uncertainties. However, in complex scenarios with occlusions and missed detections, track uncertainties can become quite large. As a result, uncertainties of predicted tracks can become very large, which eventually results in divergence of track prediction errors. In this case, deterministic track states are no longer appropriate and uncertainties should be incorporated in a proper form to reflect the inherent randomness existing in track motions.

As for detections, the effectiveness of object detectors varies with different scenes and different objects. On the one hand, for close objects that are not occluded, detections can be usually obtained with high accuracy in position and shape, as well as high detection confidence. In other words, uncertainties of detections generated by such objects are relatively low. On the other hand, for distant objects or occluded objects, detections become less reliable and the accuracy and confidence of detections are reduced accordingly, i.e. the level of detection uncertainty is increased. For the first scenario above, good data association performance can be achieved using deterministic detection states. However, for the second case, deterministic detection states cannot truly reflect real object status, which can easily lead to data association failure.

In order to overcome the above shortcomings and describe the states of tracks and detections more faithfully, we propose in this paper a TBD-based 3D MOT method "3D multi-object tracking based on uncertainty-guided data association (UG3DMOT)" In this method, we discard the commonly used deterministic tracks and deterministic detections in the data association stage, instead, we propose to model tracks and detections as random vectors for data association. The components of these random vectors are individual Gaussian random variables representing different properties of tracks and detections. On this basis, association of tracks and detections is achieved by evaluating the similarity between these two random vectors.

This work was supported by the Chongqing Technology Innovation and Application Development Project under Grant CSTB2022TIAD-DEX0013. (*Corresponding author: Chunyun Fu*)

Jiawei He and Xiyang Wang are with the College of Mechanical and Vehicle Engineering, Chongqing University, Chongqing 400044, China (e-mail: hejiawei@cqu.edu.cn; wangxiyang@cqu.edu.cn).
Chunyun Fu is with the State Key Laboratory of Mechanical Transmissions and the College of Mechanical and Vehicle Engineering, Chongqing University, Chongqing 400044, China (email: fuchunyun@cqu.edu.cn)

The proposed method was evaluated based on the commonly used KITTI dataset [12] and nuScenes dataset [13]. The results showed that our proposed tracker achieved an advanced level of accuracy compared with existing state-of-the-art methods. The main contributions of this paper can be summarized as follows: first, based on the TBD framework, we propose a novel data association mechanism for MOT by modeling tracks and detections as random vectors. By this means, the inherent uncertainties existing in tracks and detections are taken into consideration, and the conventional deterministic data association mechanism is discarded. Second, based on the Jensen-Shannon divergence [14], a new cost function is designed to evaluate the similarity between tracks and detections. With this cost function, the range for data association is expanded to compensate for the increase of uncertainty resulting from continuous track predictions. Lastly, the level of track uncertainty is employed in our proposed method to guide the data association process, which solves the 'multiple-association problem' due to the expanded association range of tracks.

## II. RELATED WORKS

### A. 3D MOT Overview

In recent years, the rapid development of 3D detection technology has accelerated the advancement of 3D MOT. In the existing literature, TBD-based 3D MOT methods usually employ 3D object information extracted from images [1], [15] or point clouds [4], [5] to realize tracking in the 3D domain. These 3D MOT methods are normally composed of four major steps: track prediction, data association, track update, and track life management [8], [16], [17]. Since the focus of this paper is on data association, the following literature review will be only concentrated on this topic.

### B. Data Association in 3D MOT

As mentioned above, the major purpose of data association is to establish one-to-one correspondence between tracks and detections. To this end, various types of cost functions, such as Intersection over Union (IoU), are proposed in the existing data association algorithms. These cost functions are used to evaluate the similarities between tracks and detections in terms of different aspects, such as appearance, geometry and motion. Weng et al. [7] proposed a 3D tracking method which implements 3D MOT by extending the concept of IoU from the 2D domain to the 3D domain. This work has become a typical baseline in the field of 3D MOT, commonly referred to as 'AB3DMOT'. Although this approach is computationally efficient, it is not robust enough for low-frame-rate scenes (such as the nuScenes dataset). The main reason is that the 3D IoU used in this method becomes non-informative when two bounding boxes do not overlap. In other words, the relative distance between two bounding boxes cannot be reflected in this IoU design if no overlap exists, since in this case the IoU value always remains zero no matter how far one bounding box is from the other. Similarly, this problem also exists in the existing 2D MOT works that employ 2D IoU as the cost function. In response to this problem, some MOT methods proposed to use GIoU [18], DIoU [19], SDIoU [20] and other variants to replace the traditional IoU, aiming to improve the robustness of data association.

Apart from the commonly used baseline — AB3DMOT [7], other important 3D MOT works have also been proposed in recent years. Yin et al. [5] proposed to represent 3D objects as points for tracking, which simplified the 3D MOT problem to a greedy closest point matching problem. Kim et al. [11] proposed a 3D MOT method based on multi-modality sensor fusion. This method employed a modified Euclidean distance as the cost function in which the orientation similarity between two bounding boxes was taken into consideration, and it was pointed out in this paper that adding a penalty term of orientation to the original Euclidean distance is beneficial for distinguishing different types of objects. Wu et al. [8] proposed to use an aggregated pairwise cost for data association, which takes into account the object similarity in terms of appearance, geometry, and motion.

The cost functions used in the above MOT methods are all based on deterministic tracks and detections, and the inherent uncertainties existing in tracks and detections are overlooked. However, the existence of uncertainties inevitably reduces the accuracy and effectiveness of data association, and even leads to association failure in extreme scenarios (e.g. consecutive frames of track predictions due to occlusion). Although some solutions have been proposed in the literature to tackle this problem, the efficacy of these remedies is limited. For example, Reich et al. [21] and Chiu et al. [16] proposed to use the Mahalanobis distance to reflect uncertainties in tracks and detections. Specifically, in these methods, an innovation covariance is obtained by simply superimposing the uncertainties of tracks and detections, which is then used to correct the Euclidean distance. In challenging scenarios where track uncertainties are increased due to continuous track prediction, the association mechanisms proposed in these two methods are prone to associate most objects in the scene, which leads to significant performance deterioration or even failure of data association. To tackle the above shortcomings, firstly, we propose to model tracks and detections as random vectors, with components being individual Gaussian random variables. Secondly, for data association purposes, we use the Jensen–Shannon divergence in conjunction with a heading penalty term to measure evaluate the similarity between tracks and detections. Then, we introduce the level of track uncertainty into the cost function, so as to guide the data association process. By means of the above three measures, the drawbacks resulting from using deterministic tracks and detections is alleviated, and the tracking performance is enhanced in the presence of inherent uncertainties.

As shown in Fig. 1, the method proposed in this paper follows the commonly used TBD framework, and it is composed of three major modules: state modeling, data association, and track management. In the first module, the object bounding boxes produced by the 3D detector are used to generate random vector representations of 3D detections.

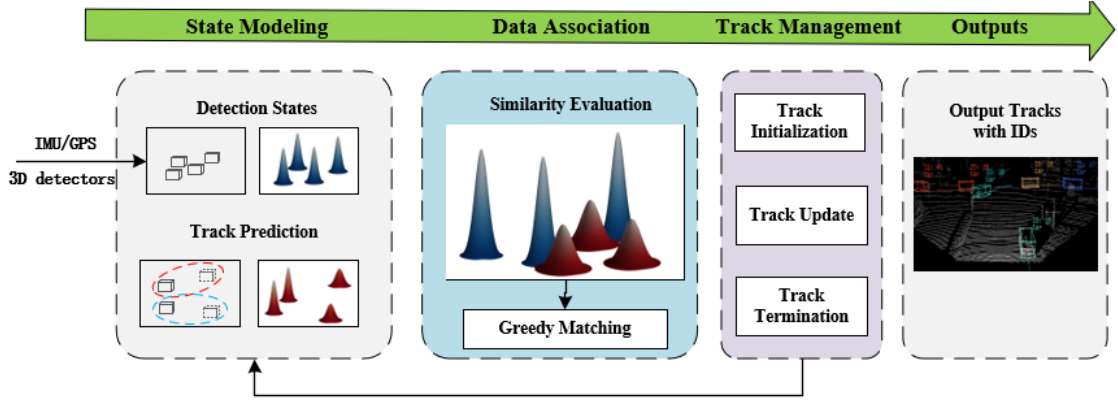

Fig. 1. Overall structure of the proposed 3D MOT framework based on uncertainty-guided data association.

## III. PROPOSED METHOD

Likewise, 3D tracks are also represented by random vectors in our approach, by means of Kalman filters. In the second module, we no longer employ deterministic tracks and deterministic detections for similarity evaluation. Instead, we propose to use the JS divergence with a heading penalty term to measure the similarity between two multidimensional Gaussian distributions (i.e. track and detection), and then we introduce track uncertainty into the cost function to eliminate the 'multiple-association problem'. The third module, track management, is mainly used to manage the track lifecycle. A typical track management strategy in the literature is used in our method. The three modules of our proposed MOT method are described in detail below.

### A. State Modeling

Appropriate modeling of track states and detection states significantly affects the performance of a tracker. In this study, tracks and detections are modeled as random vectors whose components are individual Gaussian random variables. By this means, uncertainties are introduced in the states of tracks and detections, which lays the foundation of the entire MOT framework.

The current 3D object detectors can obtain various information regarding the detected objects, including position, dimension, orientation, and detection confidence. Note that the original detections generated by detectors are with respect to the sensor coordinate system. To accomplish 3D tracking in the world coordinate system, the detections expressed in the sensor coordinate system are transformed to the world coordinate system, based on the GPS/IMU measurements. In this study, we express the transformed 3D detections in the world coordinate system as follows:

$$\{D_t^i = [{}^D x_t^i, {}^D y_t^i, {}^D z_t^i, {}^D \theta_t^i, {}^D l_t^i, {}^D w_t^i, {}^D h_t^i]^T\} \quad (1)$$

where $D_t^i$ denotes the $i$-th detection at time $t$, ${}^D x_t^i$, ${}^D y_t^i$ and ${}^D z_t^i$ represent the coordinates of $D_t^i$ in the 3D domain, ${}^D \theta_t^i$ stands for heading angle of $D_t^i$, and ${}^D l_t^i$, ${}^D w_t^i$, and ${}^D h_t^i$ indicate the length, width and height of $D_t^i$.

As mentioned above, in this study the components of $D_t^i$ are modeled as individual Gaussian random variables. To reflect the variations of these random variables, their covariance matrix is defined as follows:

$$\Sigma_t^i = \text{diag}(\sigma_x^2, \sigma_y^2, \sigma_z^2, \sigma_\theta^2, \sigma_l^2, \sigma_w^2, \sigma_h^2) \quad (2)$$

where $\sigma_x^2, \sigma_y^2, \sigma_z^2, \sigma_\theta^2, \sigma_l^2, \sigma_w^2$ and $\sigma_h^2$ represent the variances of $x, y, z, \theta, l, w$ and $h$, respectively. Note that the covariance matrix is a diagonal matrix as it is assumed in this study that the seven components of $D_t^i$ are mutually independent Gaussian random variables.

To sum up, the $i$-th detection obtained by the 3D detector at time $t$, $D_t^i$ can be modeled as a random vector as follows:

$$D_t^i \sim N(\mu_t^i, \Sigma_t^i) \quad (3)$$
$$\mu_t^i = E[D_t^i] \quad (4)$$

where $\mu_t^i$ denotes the mean vector of $D_t^i$.

*2) Track state modeling:* Similar to detections, in this study, tracks are also modeled as random vectors with Gaussian random variables being components. Specifically, tracks can be expressed in the following form:

$$\{T_t^j = [{}^T x_t^j, {}^T y_t^j, {}^T z_t^j, {}^T \theta_t^j, {}^T l_t^j, {}^T w_t^j, {}^T h_t^j, {}^T \dot{x}_t^j, {}^T \dot{y}_t^j, {}^T \dot{z}_t^j]^T\} \quad (5)$$

where $T_t^j$ represents the $j$-th track at time $t$, ${}^T x_t^j$, ${}^T y_t^j$ and ${}^T z_t^j$ denote the coordinates of $T_t^j$ in the 3D domain, ${}^T \theta_t^j$ stands for heading angle of $T_t^j$, and ${}^T l_t^j$, ${}^T w_t^j$, and ${}^T h_t^j$ indicate the length, width and height of $T_t^j$. Note that in this study a constant velocity (CV) model [7], [22] is employed for object speed propagation in consecutive frames, and the covariance matrix for $T_t^j$ is obtained through a Kalman filter.

In the tracking process, we commonly encounter situations where 3D detections fail to be associated with existing tracks. In this case, each unmatched detection will then be initialized as a new track, which is assigned an initial state vector $T_t^j$ and an initial covariance matrix $\Sigma_t^j$. Specifically, the states of an unmatched detection $D_t^i$, including $x_t^i, y_t^i, z_t^i, \theta_t^i, l_t^i, w_t^i$ and $h_t^i$, are initialized as the first seven components of the new track $T_t^j$. Then, the last three velocity components of $T_t^j$ are initialized as zero since they cannot be directly obtained from detection $D_t^i$. As for the covariance matrix of this new track, its first seven diagonal components are the same as $\Sigma_t^i$, and the last three variances (denoted by $\Sigma_{\text{kin}}$) correspond to the last three Gaussian velocity components in $T_t^j$. Hence, the complete covariance matrix of the initialized new track takes the following form:

$$\Sigma_t^j = \begin{pmatrix} \Sigma_t^i & \\ & \Sigma_{\text{kin}} \end{pmatrix} \in \mathbb{R}^{10\times 10} \tag{6}$$

where $\Sigma_{\text{kin}} = \text{diag}(\sigma_{\dot{x}}^2, \sigma_{\dot{y}}^2, \sigma_{\dot{z}}^2)$.

In the existing tracking frameworks, prediction of track states is commonly conducted to propagate track states from one frame to the next. In this study, we follow the standard Kalman filter prediction process, based on the CV model and the assumption of zero-mean Gaussian process noise. The Kalman filter prediction equations for the $j$-th track are given as follows:

$$\hat{T}_t^j = A T_{t-1}^j \tag{7}$$
$$\hat{\Sigma}_t^j = A \Sigma_{t-1}^j A^{\text{T}} + Q \tag{8}$$

where $\hat{T}_t^j$ represents the prediction of the $j$-th track at time $t$, $\hat{\Sigma}_t^j$ denotes the predicted covariance matrix of the $j$-th track at time $t$, $A$ stands for the state transition matrix based on the CV model, and $Q$ indicates the covariance matrix of the process noise. In the following section, how matrix $Q$ can be estimated will be clearly explained.

In the TBD-based tracking paradigm, the process of data association leads to successfully matched detection-track pairs. For these matched tracks, their states are updated using detections produced by respective 2D or 3D detectors, thereby reducing the extent of track uncertainties. In this study, we follow the standard Kalman filter update process for the $j$-th track, as follows:

$$K_t = \hat{\Sigma}_t^j H^{\text{T}} (H \hat{\Sigma}_t^j H^{\text{T}} + R)^{-1} \tag{9}$$
$$T_t^j = \hat{T}_t^j + K_t (D_t^i - H \hat{T}_t^j) \tag{10}$$
$$\Sigma_t^j = (I - K_t H) \hat{\Sigma}_t^j \tag{11}$$

where $R$ represents the covariance matrix of the measurement noise, $H$ denotes observation model, $D_t^i$ is the detection which is successfully associated with $T_t^j$. Note that in this study the measurement noise is assumed to be zero-mean Gaussian noise, and the estimation of matrix $R$ is elaborated in the following section.

*3) Estimation of uncertainties:* The major uncertainties involved in this study, such as covariance matrices $Q$ and $R$, can be estimated in an intuitive manner using the method given in [7]. However, this approach becomes inappropriate if uncertainties are involved in the computation of cost functions for data association, which can lead to deterioration of tracking performance. Besides, Reich and Wuensche [21] specified measurement noise and process noise considering various aspects such as object distances from the sensor. However, since this method is still a heuristic approach, its effectiveness is also limited.

In this study, the uncertainties involved in detections and tracks are estimated using the method proposed in [22]. Instead of constructing covariance matrices in an intuitive or heuristic fashion, Chiu et al. [22] proposed to estimate the covariance matrices using the statistics of the training set data. By this means, the covariance matrices $\Sigma_{\text{kin}}$, $Q$ and $R$ in this paper can be determined.

*B. Data Association*

Data association plays a crucial role in the TBD-based tracking paradigm. In the existing MOT works, most data association strategies rely on computation of cost functions (such as 3D IoU and its variants, and Euclidean distance between bounding boxes) in which deterministic states of tracks and detections are used as inputs. In these works, the randomness and uncertainties naturally existing in tracks and detections have not been taken into consideration, which reduces the effectiveness and robustness of data association. To tackle this limitation, in this study, we carefully take into account randomness and uncertainties by modeling detections and tracks as random vectors. Then, we employ Jensen-Shannon divergence [14] to evaluate the similarity between tracks and detections, and complete data association by means of the greedy match algorithm. The detailed data association process is explained as follows.

*1) Cost function:* The Jensen-Shannon (JS) divergence, also known as information radius [23] or total divergence to the average [24], is a commonly used indicator in statistics to evaluate the similarity between two distributions. Although the JS divergence is based on the Kullback-Leibler (KL) divergence [25], it is different from the KL divergence in several aspects. For instance, the JS divergence is symmetric while the KL divergence is asymmetric. Assuming two distributions $p(x)$ and $q(x)$, their KL divergence and JS divergence are expressed as [14], [25]:

$$D_{\text{KL}}(p||q) = \int_x p(x) \log \frac{p(x)}{q(x)} \tag{12}$$
$$D_{\text{JS}}(p||q) = \frac{1}{2} D_{\text{KL}}(p||m) + \frac{1}{2} D_{\text{KL}}(q||m) \tag{13}$$
$$m = \frac{1}{2}(p + q) \tag{14}$$

where $D_{\text{KL}}(p||q)$ denotes the KL divergence, $D_{\text{JS}}(p||q)$ represents the JS divergence. It can be readily seen from (12), (13) that the KL divergence is asymmetric (i.e. $D_{\text{KL}}(p||q) \neq D_{\text{KL}}(q||p)$), while the JS divergence is symmetric (i.e. $D_{\text{JS}}(p||q) = D_{\text{JS}}(q||p)$) due to the introduction of distribution $m$.

Now, considering two $k$ dimensional Gaussian distributions $p \sim N(\mu_p, \Sigma_p)$ and $q \sim N(\mu_q, \Sigma_q)$, their KL divergence can be expressed as follows [26]:

$$D_{\text{KL}}(p||q) = \frac{1}{2}\left[\log \frac{|\Sigma_q|}{|\Sigma_p|} - k + (\mu_p - \mu_q)^{\text{T}} \Sigma_q^{-1} (\mu_p - \mu_q) + \text{tr}\{\Sigma_q^{-1} \Sigma_p\}\right] \tag{15}$$

where $\Sigma_p$ denotes the covariance matrix of $p$, $\Sigma_q$ represents the covariance matrix of $q$, $\mu_p$ indicates the mean vector of $p$, and $\mu_q$ stands for the mean vector of $q$. Besides, based on (12), the corresponding expression of JS divergence for two Gaussian distributions can be readily derived.

In this paper, we do not directly employ the original form of JS divergence $D_{\text{JS}}$ to evaluate the similarity between detections and tracks. Instead, $D_{\text{JS}}$ is multiplied by a penalty term which reflects the bearing of an object, in order to include the influence of object bearing. The underlying reason for this modification is that large uncertainties in the object bearing can lead to large covariances accordingly. This modified divergence, denoted by $D_{\text{mod}}$, is given as follows:

$$D_{\text{mod}}(D_t^i || \hat{T}_t^j) = D_{\text{JS}}(D_t^i || \hat{T}_t^j) \times \alpha(D_t^i, \hat{T}_t^j) \tag{16}$$
$$\alpha(D_t^i, \hat{T}_t^j) = 2 - \cos\langle \theta_{D_t^i}, \theta_{\hat{T}_t^j}\rangle \in [1,2] \tag{17}$$

where $\theta_{D_t^i}$ denotes the heading angle of detection $D_t^i$, and $\theta_{\hat{T}_t^j}$ represents the heading angle of track $\hat{T}_t^j$. It is seen from equations (16)-(17) that a large discrepancy between the headings of $D_t^i$ and $\hat{T}_t^j$ leads to a large $\alpha(D_t^i, \hat{T}_t^j)$, thereby increasing the value of $D_{\text{mod}}(D_t^i||\hat{T}_t^j)$ and reducing the association possibility of $D_t^i$ and $\hat{T}_t^j$.

Apart from the introduction of the penalty term for heading discrepancy, another important feature of this cost function is: the range of track-detection association expands as the number of track prediction increases. We might as well elaborate on this point using Fig. 2. In this example, we see that as prediction frames increase, the mean covariance [1] of the predicted track increases accordingly, and the resulting cost $D_{\text{mod}}(D_t^i||\hat{T}_t^j)$ drops. With this cost function, for the same detection, the tracks that are located farther away can be involved in association, in order to compensate for the increase of uncertainty due to consecutive frames of track prediction. This feature of cost function indicates that when dealing with missed detections in challenging scenarios (e.g. occlusions and sensor malfunctions), our approach is able to quickly resume association due to the reduced cost value.

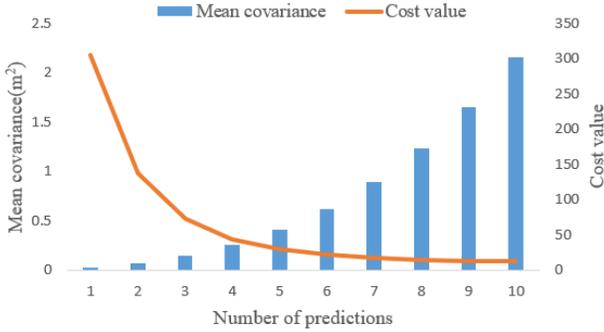

Fig. 2. Mean covariance and cost value with increasing number of track predictions. In this approach, track prediction is realized using the standard Kalman filter prediction process, with which the mean covariance of the predicted track increases while the cost value (for the same detection) drops, as the number of track predictions increases.

*2) Uncertainty-guided data association:* As mentioned previously, using the proposed cost function can facilitate data association by expanding the range for association, which is especially beneficial for dealing with challenging scenarios (e.g. occlusions and sensor malfunctions). However, this benefit can possibly lead to problems such as "multiple-association", as demonstrated in Fig. 3. The underlying reason for these problems lies in the increasing level of track uncertainty caused by continuous track prediction.

---

[1] Mean covariance: we define the mean covariance of a predicted track as the average of location-related and dimension-related elements in the covariance matrix. Specifically, it can be expressed as: $\text{mean}(\hat{\Sigma}_t^j) = \frac{(\sigma_x^2+\sigma_y^2+\sigma_z^2+\sigma_l^2+\sigma_w^2+\sigma_h^2)}{6}$. Note that the heading-related element $\sigma_\theta^2$ in the covariance matrix is not involved in $\text{mean}(\hat{\Sigma}_t^j)$, because its effect has already been taken into account in the penalty term $\alpha(D_t^i, \hat{T}_t^j)$.

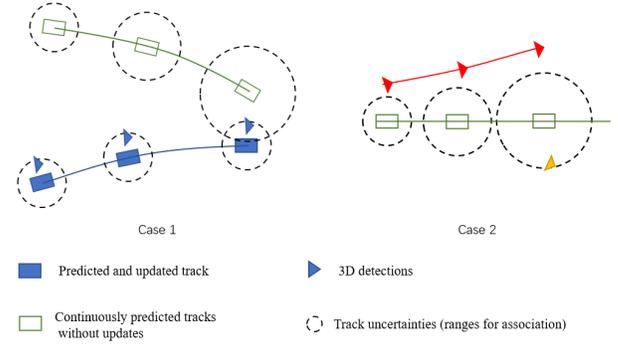

Fig. 3. Typical problems resulting from continuous prediction of track states. Case 1 shows that in challenging scenarios (such as occlusion), consecutive frames of track predictions inevitably increase the uncertainty of the predicted track. As a result, using the proposed cost function, the detection can possibly be associated with more than one track – causing the 'multiple-association problem'. In Case 2, the red detections are initialized as a new track, because they cannot be associated with any tracks. However, the yellow detection, which would be initialized as a new track if no continuous track prediction occurred, is instead associated with the predicted track because the uncertainty of this predicted track increases as prediction continues, which makes the range for association expands accordingly.

To prevent the problems presented in Fig. 3 from happening, inspired by [8], in this paper we introduce the track covariance into the cost function to incorporate the effect of track uncertainty on data association. The resulting cost function, denoted by $D_{\text{res}}(D_t^i||\hat{T}_t^j)$, is then written as:

$$D_{\text{res}}(D_t^i||\hat{T}_t^j) = D_{\text{mod}}(D_t^i||\hat{T}_t^j) \times \text{mean}(\hat{\Sigma}_t^j) \quad (18)$$

where $\text{mean}(\hat{\Sigma}_t^j)$ denotes the mean covariance of the predicted track as described in footnote 1.

### C. Track Management

Track management plays an important role in most existing MOT works. It has been shown in [7], [17] that a well-designed track management strategy is conducive to suppressing false positives (FPs) and false negatives (FNs). In this study, the track management strategy proposed in [7] is employed to handle objects entering or exiting the scene. This strategy relies on properly designed birth and death memories, by which means tracks can be categorized into three types: confirmed, tentative, and disappeared. Interested readers are referred to [7] for more details about this track management strategy.

## IV. EXPERIMENTS

### A. Datasets, Baselines and Evaluation Metrics

**Datasets:** The commonly used KITTI and nuScenes datasets are employed in this study for evaluation purposes. It should be pointed out that the CenterPoint 3D detector [5] is used for the nuScenes dataset and the CasA 3D detector [4] is used for the KITTI dataset, for all competing MOT methods discussed in this paper.

**Baseline Methods:** To evaluate the effectiveness of the proposed approach, the state-of-the-art MOT works on the leader borads are employed for comparative studies, including

PermaTrack [27], PC3T [8], Mono-3D-KF [21], TripletTrack [28], DeepFusionMOT [9], PolarMOT [29], DiTMOT [30], SimpleTrack [17], Belief-Propagation [31], IPRL-TRI-M [16], CenterPoint [5] and JMODT [32].

**Evaluation Metrics:** In this paper, the evaluation metrics used by the KITTI and nuScenes datasets are employed to evaluate the tracking performance of the above competing methods. These commonly used metrics include, but are not limited to, CLEAR [33], HOTA [34] and AMOTA [7]. The CLEAR metric is composed of two sub-metrics, multi-object tracking accuracy (MOTA) and multi-object tracking precision (MOTP). The HOTA metric is a major evaluation metric used by the KITTI dataset, while the AMOTA is a major evaluation metric used by the nuScenes dataset.

## B. Experiment Results

**Quantitative Results:** Comparative results using the KITTI dataset are given in Table I for eight competing tracking methods. We see that in terms of the key evaluation metric, HOTA, our proposed method achieved the best result among all eight methods. Besides, as for another important metric, ID switch (IDSW), our method also provided the best value. Note that this IDSW number, 30, is far lower than the results of any other compared methods. This significant advantage indicates that the proposed method provides accurate data association and maintains stable tracking in challenging scenarios such as occlusions and missed detections.

Apart from the KITTI dataset, quantitative results using the nuScenes dataset are demonstrated in Table II. This table shows the performances of the six competing methods in terms of the key evaluation metric, AMOTA, for the nuScenes test sequences. We see that the proposed method achieved the best overall AMOTA value, along with several other classes including bus, trailer and truck. It is known that the AMOTA metric reflects the overall tracking performance of an MOT method in terms of FP, FN and IDSW over all recall values [7]. Hence, the best AMOTA value attained by the proposed method indicates that the proposed method outperforms its competitors with an improved all-round 3D tracking performance.

**Qualitative Results:** Apart from quantitative results, we also demonstrate qualitative results to visualize the trajectory of an object in the tracking process. Specifically, the scenes in frames 93—162 in training sequence 0002 of the KITTI dataset were used in our visualization. These scenes demonstrate an urban intersection where multiple occlusions existed. In Fig. 4(a), we see that the uncertainties of the tracked object varied from time to time. Each surge of uncertainty was caused by an occurrence of occlusion. For example, in Fig. 4(b), we see that the vehicle encircled by the red bounding box started being occluded since frame 99, and then reappeared in the sensor FOV and recaptured by the detector in frame 102. The uncertainty of this vehicle significantly increased during the frames in which it was occluded. Note that as time elapses, the object being tracked is getting increasingly far from the sensor and the detector eventually fails to produce any detection, which in turn leads to divergence of the uncertainty curve in Fig. 4(a). With the proposed MOT method, we obtain the trajectory of the tracked object, as shown in Fig. 4(b). We see from Fig. 4(b) that when dealing with highly uncertain situations, i.e. occlusions in our experiments, the proposed method maintained stable tracking of the object by means of the proposed uncertainty-guided data association approach.

TABLE I
TRACKING PERFORMANCE COMPARISON USING THE KITTI TEST SEQUENCES ('CAR' CLASS)

| Method | Published Year | HOTA (%)↑ | MOTA (%)↑ | IDSW ↓ |
|---|---|---|---|---|
| PC3T [8] | TITS (2021) | 77.80 | **88.81** | 225 |
| Mono-3D-KF [21] | IEEE Fusion (2021) | 75.47 | 88.48 | 162 |
| DeepFusionMOT [9] | RAL (2022) | 75.46 | 84.63 | 84 |
| PolarMOT [29] | ECCV (2022) | 75.16 | 85.08 | 462 |
| TripletTrack [28] | CVPR (2022) | 73.58 | 84.32 | 322 |
| DiTMOT [30] | RAL (2021) | 72.21 | 84.53 | 101 |
| JMODT [32] | IROS (2021) | 70.73 | 77.39 | 350 |
| Our Method | -- | **78.6** | 87.98 | **30** |

TABLE II
TRACKING PERFORMANCE COMPARISON USING THE NUSCENES TEST SET IN TERMS OF AMOTA.

| Method | Published Year | Overall↑ | Bicycle↑ | Bus↑ | Car↑ | Motorcycle↑ | Pedestrian↑ | Trailer↑ | Truck↑ |
|---|---|---|---|---|---|---|---|---|---|
| SimpleTrack [17] | arXiv (2021) | **0.668** | 0.407 | 0.715 | 0.823 | **0.674** | 0.796 | 0.673 | 0.587 |
| Belief Propagation-V2 [31] | Proc. of the IEEE (2022) | 0.666 | 0.397 | 0.703 | 0.837 | 0.64 | 0.8 | 0.689 | 0.599 |
| PolarMOT [29] | ECCV (2022) | 0.664 | 0.349 | 0.708 | **0.853** | 0.656 | **0.806** | 0.673 | 0.602 |
| IPRL-TRI-M [16] | ICRA (2021) | 0.655 | **0.469** | 0.713 | 0.83 | 0.631 | 0.741 | 0.657 | 0.546 |
| CenterPoint [5] | CVPR (2021) | 0.650 | 0.331 | 0.715 | 0.818 | 0.587 | 0.78 | 0.693 | 0.625 |
| Our Method | -- | **0.668** | 0.366 | **0.739** | 0.817 | 0.660 | 0.773 | **0.697** | **0.625** |

TABLE III
PERFORMANCE COMPARISON OF TWO COST FUNCTIONS ON THE KITTI TRAINING SEQUENCES, IN TERMS OF DIFFERENT EVALUATION METRICS.

| Cost function | HOTA↑ | DetA↑ | AssA↑ | MOTA↑ | FN ↓ | FP ↓ | IDSW ↓ |
|---|---|---|---|---|---|---|---|
| Mahalanobis distance | 80.092 | 77.355 | 83.187 | 85.692 | 2342 | 1081 | 21 |
| Proposed cost function | **80.879** | **78.221** | **83.885** | **86.67** | **2221** | **963** | **14** |

TABLE IV
PERFORMANCE COMPARISON OF TWO COST FUNCTIONS ON THE NUSCENES VALIDATION SET, IN TERMS OF OVERALL AMOTA AND INDIVIDUAL AMOTA FOR VARIOUS OBJECT CATEGORIES.

| Cost function | Overall↑ | Bicycle↑ | Bus↑ | Car↑ | Motorcycle↑ | Pedestrian↑ | Trailer↑ | Truck↑ |
|---|---|---|---|---|---|---|---|---|
| Mahalanobis distance | 0.672 | 0.477 | 0.851 | 0.775 | 0.645 | 0.811 | 0.485 | 0.663 |
| Proposed cost function | **0.698** | **0.506** | **0.855** | **0.827** | **0.682** | **0.816** | **0.51** | **0.689** |

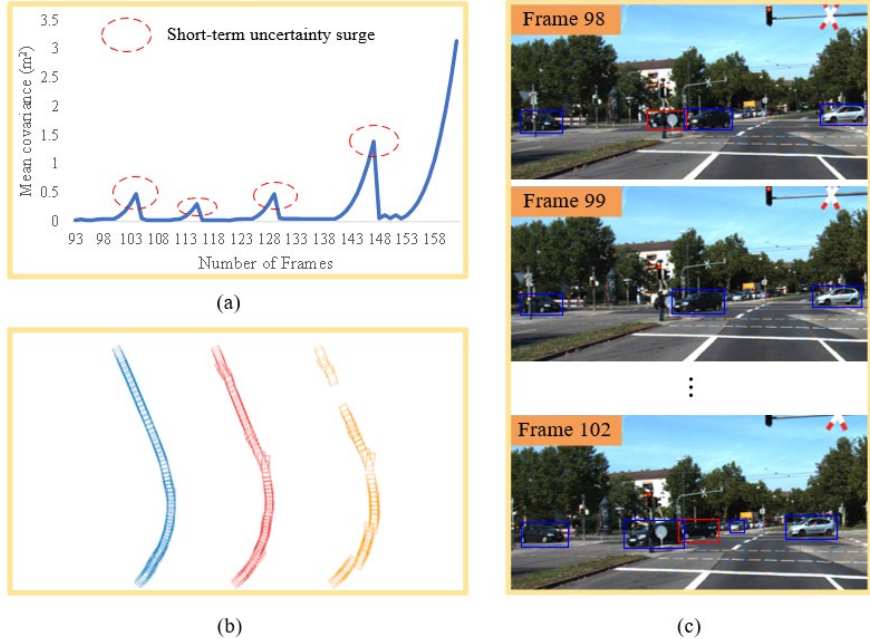

Fig. 4. Visualization of an object trajectory in the tracking process. (a) shows how the level of uncertainty evolves, (b) shows the ground truths (blue), the sensor detection in each frame (orange), and the tracked object in each frame (red), and (c) demonstrates an instance of occlusion in the tracking process.

**Ablation Study:** As mentioned previously, the main contributions of this paper lie in the modelling of tracks and detections as random vectors which reflect their inherent uncertainties, as well as the proposal of a new cost function to achieve uncertainty-guided data association. In this section, we first compare our proposed cost function with another typical cost function design – Mahalanobis distance, and then investigate the effects of the uncertainty-guided data association mechanism on the tracking performance.

Cost function design: Cost function plays a crucial role in most current MOT methods. The tracking performances of our proposed method and the Mahalanobis distance are compared in Tables III and IV. As to the KITTI dataset, it is seen in Table III that our proposed cost function leads to superior tracking results, including HOTA, DetA, AssA, MOTA, FP and IDSW. As for the nuScenes dataset, we see in Table IV that the proposed cost function outperforms the Mahalanobis distance, in terms of the overall AMOTA and individual AMOTA for various object categories. Note that in this ablation study, except the cost function, all other components of the tracker remained the same.

Uncertainty-guided data association: Tables V and VI show that with the proposed uncertainty-guided data association mechanism, the tracking performance is enhanced for both KITTI and nuScenes datasets. Besides, the reductions of FN, FP and IDSW in this experiment indicate that the problems (e.g. multiple-association) caused by continuous track prediction were effectively alleviated by the proposed uncertainty-guided data association mechanism.

TABLE V
COMPARISON RESULTS ON THE KITTI TRAINING SEQUENCES USING DIFFERENT DATA ASSOCIATION (DA)

| DA mechanism | HOTA ↑ | MOTA ↑ | FN ↓ | FP ↓ | IDSW ↓ |
|---|---|---|---|---|---|
| No uncertainty involved | 79.751 | 85.422 | 2395 | 1075 | 34 |
| Uncertainty-guided | **80.879** | **86.67** | **2231** | **963** | **14** |

TABLE VI
COMPARISON RESULTS ON THE NUSCENES TRAINING SET USING DIFFERENT DATA ASSOCIATION MECHANISMS.

| DA mechanism | AMOTA (%)↑ | Recall (%)↑ | FN ↓ | FP ↓ | IDSW ↓ |
|---|---|---|---|---|---|
| No uncertainty involved | 0.662 | 0.700 | 23050 | 14134 | 693 |
| Uncertainty-guided | **0.698** | **0.740** | **19200** | **13542** | **655** |

## V. CONCLUSION

In the existing literature, most MOT methods rely on deterministic models for evaluating similarities between tracks and detections, and do not take into account the inherent uncertainties of tracks and detections. In this paper, we proposed to model tracks and detections as random vectors which reflect their inherent uncertainties. On this basis, we designed a JS divergence-based cost function to evaluate the similarities between tracks and detections. To tackle the problems caused by consecutive frames of track predictions, track uncertainty was incorporated into the cost function design, thereby achieving uncertainty-guided data association. Experiments were conducted on both KITTI and nuScenes datasets, and the proposed method showed superior tracking performance in comparison with several state-of-the-art MOT methods.